\documentclass[final]{cvpr}

\usepackage{times}
\usepackage{epsfig}
\usepackage{graphicx}
\usepackage{amsmath}
\usepackage{amssymb}

\usepackage{enumitem} 
\usepackage{amsthm}
\newtheorem{theorem}{Theorem}

\usepackage[
  separate-uncertainty = true,
  multi-part-units = repeat
]{siunitx}

\usepackage{hyperref}
\hypersetup{pagebackref=true,breaklinks=true,colorlinks,bookmarks=false}

\newcommand{\trinity}{{\it TrinityAI}}

\def\cvprPaperID{2758} 
\def\confYear{CVPR 2021}

\begin{document}

\title{Detecting Trojaned DNNs Using Counterfactual Attributions}

\author{
Karan Sikka* \;\; Indranil Sur* \;\;  Susmit Jha \;\;  Anirban Roy \;\;  Ajay Divakaran\\
SRI International\\
{\tt\small \{karan.sikka,indranil.sur, susmit.jha, anirban.roy, ajay.divakaran\}@sri.com} \\
{\tt\small *denotes equal contribution}
}

\maketitle

\def\ie{\textit{i.e.}}
\def\eg{\textit{e.g.}}

\definecolor{redcol}{rgb}{1, 0, 0}
\definecolor{bluecol}{rgb}{0, 0, 1}
\newcommand{\red}[1]{\textcolor{redcol}{#1}} 
\newcommand{\blue}[1]{\textcolor{bluecol}{#1}} 
\renewcommand{\paragraph}[1]{\smallskip\noindent{\bf{#1}}}
\newcommand{\todo}[1]{\red{TODO: {#1}}}
\newcommand{\colons}[1]{``{#1}''}

\def\algorithmautorefname{Algorithm}
\def\figureautorefname{Figure}
\def\tableautorefname{Table}
\def\equationautorefname{Eq.}
\def\sectionautorefname{Section}

\begin{abstract}
We target the problem of detecting Trojans or backdoors in DNNs. Such models behave normally with typical inputs but produce specific incorrect predictions for inputs poisoned with a Trojan trigger. Our approach is based on a novel observation that the trigger behavior depends on a few ghost neurons that activate on trigger pattern and exhibit abnormally higher relative attribution for wrong decisions when activated. Further, these trigger neurons are also active on normal inputs of the target class. Thus, we use counterfactual attributions to localize these ghost neurons from clean inputs and then incrementally excite them to observe changes in the model's accuracy. We use this information for Trojan detection by using a deep set encoder that enables invariance to the number of model classes, architecture, etc. Our approach is implemented in the \trinity\;  tool that exploits the synergies between trustworthiness, resilience, and interpretability challenges in deep learning. We evaluate our approach on benchmarks with high diversity in model architectures, triggers, etc. We show consistent gains ($+10\%$) over state-of-the-art methods that rely on the susceptibility of the DNN to specific adversarial attacks, which in turn requires strong assumptions on the nature of the Trojan attack. 

\end{abstract}

\section{Introduction}
\label{sec:intro}

Deep neural networks (DNNs) have emerged as the representation of choice for machine learning models in multiple domains \cite{krizhevsky2012imagenet, bahdanau2014neural, graves2013speech}.  
The ease of training large-scale DNNs with huge amounts of data has made it possible to achieve
near human-level performance on several benchmarks. 
Despite these successes, DNNs are known to be
fragile and vulnerable to adversarial attacks which inhibits their adoption in high-assurance safety-critical systems such as autonomous driving and user authentication.  The space of adversarial attacks on DNNs is diverse ranging from inference-time adversarial perturbations that lead to incorrect prediction by the ML model~\cite{szegedy2013intriguing}, reverse-engineering attacks to infer weights of a trained DNN~\cite{shokri2015privacy}, and  training-time attacks that poison the dataset \cite{li2020backdoor}. 
These attacks and corresponding defense mechanisms for DNNs have received significant attention in  literature~\cite{carlini2019evaluating,hao2020adversarial}.  

\begin{figure}[t!]
  \centering
  \includegraphics[width=\columnwidth]{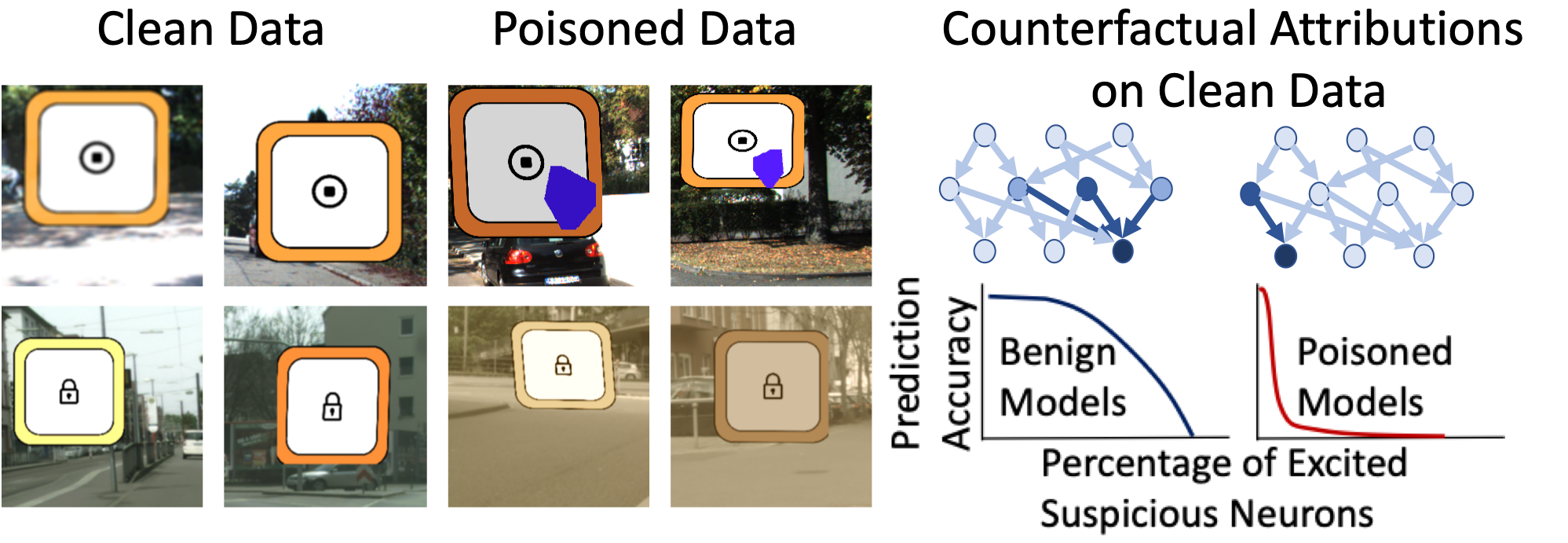}
  \caption{
  Showing clean inputs (first two columns) and Trojaned inputs (last two columns) that are poisoned with polygon and filter trigger respectively. We use counterfactual attributions over input features, which computes attributions for counter-classes that are not the actual predicted class. 
  The attributions corresponding to the Trojan trigger are localized over a few {\it ghost} neurons encoding the trigger. Exciting these suspicious ghost neurons causes the prediction accuracy of poisoned model to drop sharply compared to benign models. Our method exploits this change in accuracy to detect Trojan models.}
	\label{image:example}
  \end{figure}

Recent work~\cite{gu2017badnets,chen2017targeted, gu2017badnets} has demonstrated a new kind of training-time vulnerability where a DNN can be trained with poisoned data to be \textbf{Trojaned}. A Trojaned DNN behaves normally with high accuracy on typical inputs but can be made to produce specific incorrect predictions when the inputs contain the Trojan trigger. This paper (see \autoref{image:example}) focuses on devising a verification defense against such Trojan (also called {\it backdoor}) attacks. We develop a principled approach to verify if a trained DNN has been Trojaned, with access only to the trained model and a few clean labeled test samples. Our proposed approach builds on recent progress in explaining decisions of DNNs, and thus, draws a  connection between the interpretability of a DNN and its robustness and resilience to attacks.

A variety of triggers has been considered in Trojan attacks. 
While initial work considered stamp-like triggers~\cite{gu2017badnets},
invisible triggers have been explored by blending triggers with 
benign samples~\cite{chen2017targeted}.
Another approach to produce stealthy trigger is to perturb the benign 
sample by a backdoor trigger amplitude instead of using a stamp-like patch to replace the sample pixels~\cite{turner2019label}. 
We evaluate our method using the US IARPA/NIST-TrojAI Datasets\footnote{\url{https://pages.nist.gov/trojai/docs/data.html}} which includes a diverse set of Trojaned models with a variety of triggers, and thus requires generalizable Trojan detection approaches beyond the existing state of the art.

   Trojaned models have been known to have shortcuts in the feature space~\cite{wang2019neural,chen2019deepinspect, zhang2020cassandra} that allow triggers to switch the model's prediction from other classes to the target class. Existing detection approaches rely on indirect statistical signatures in the form of sensitivity to instance-specific or universal adversarial attacks
   \cite{zhang2020cassandra, wang2020practical}. A key limitation with such approaches is in their reliance on adversarial probes in activating the shortcut. 
  In contrast, we use attribution techniques developed for explaining DNN predictions to directly detect the presence of these {\it shortcut pathways} in the feature space \cite{sundararajan2017axiomatic,jha2019attribution, ancona2017towards}. 
 As illustrated in \autoref{image:example}, 
 we use counterfactual attributions across clean samples to obtain feature contributions for predicting counter-classes, which are different from the highest scoring (predicted) class \cite{hendricks2018generating}. 
 We mathematically show that the attributions encoding the trigger are concentrated over a few {\it ghost} neurons. 
 We exploit this property by first incrementally exciting these neurons to monitor changes in model's accuracy for each class and use these with a deep set encoder for Trojan detection 
 

\noindent The central contributions of the paper are as follows:
\begin{itemize}[noitemsep,leftmargin=*]
    \item  We are the first to observe that the Trojan triggers use \colons{poly-semantic} neurons that not only have high attribution for poisoned inputs but also show significant attribution towards target class even for clean inputs. 
    \item Using our observation of poly-semantic neurons, we develop a counterfactual attribution-based approach to detect whether a DNN  is Trojaned without access to any poisoned input or knowledge about the trigger, and with a few clean inputs. Counterfactual attributions localize \textit{ghost} neurons for each target class and we incrementally excite these to observe changes in model's accuracy.
    \item We  propose a deep temporal set encoder to make our Trojan detector invariant to the number and ordering of classes in different DNNs. This makes our approach robust to variations in the number of output classes, model architecture, trigger strength \etc. 
    \item We evaluate our approach on Trojan detection  benchmarks and demonstrate $~10\%$ improvement over the state-of-the-art approaches. Our datasets contain over $1000$ models trained on different datasets, and with high diversity in the number of classes, number of poisoned classes and the nature, shape and size of Trojan triggers. 
\end{itemize}

\section{Background and Related Work}


In the rest of the paper, we refer to the DNN with the embedded Trojan as {\it Trojaned} model, and the model without Trojan as {\it  benign} model. Training the Trojaned model is often achieved by poisoning a small fraction of the training data with inputs having the trigger pattern. These training samples with triggers are called {\it poisoned} samples. The inference time inputs with triggers are also called poisoned samples, and those without triggers are called {\it clean} samples. The original expected output of a poisoned input is called  the {\it source} class and the output of the Trojaned model is called the {\it target} class. 
The threat of Trojan attacks~\cite{gu2017badnets,chen2017targeted} on DNNs aims at embedding hidden triggers such that the poisoned inputs make the Trojaned DNN mispredict and output the target class even though its accuracy on the clean inputs remains high. 

This paper combines the fields of explaining decisions of DNNs and their adversarial robustness. We briefly discuss the related work and compare our  approach to existing techniques for detecting Trojaned models.  

\subsection{Explainability and Resilience}  
 
A number of explanation  techniques~\cite{lundberg2017unified,sundararajan2017axiomatic,li2015visual,yi2016lift,jha2017nfm,jha2019jar} 
have been recently proposed that find qualitative explanations or 
assign quantitative attributions to input features for a given  decision.  Many of these methods are based on the gradient  with respect to the 
input~\cite{simonyan2013deep,selvaraju2017grad,sundararajan2017axiomatic,adebayo2018sanity}.
A few recent theoretical studies~\cite{chalasani2018adversarial} indicate a strong  connection between the robustness of DNNs and their interpretability using attribution methods. The connection between these methods for explaining DNN decisions and detection of out-of-distribution data and adversarial examples has been related to anti-causal direction of learning~\cite{kilbertus2018generalization,jha2019safeml,jha2019attribution}. This paper is the first work to draw a connection between these explanation methods and detecting Trojaned models. 

\subsection{Adversarial Trojan Attacks and Defenses}  
Inference-time adversarial attacks and defenses for these attacks have received a lot of interest~\cite{hao2020adversarial, kurakin2018adversarial}. While adversarial perturbations are often input 
specific, universal perturbations~\cite{moosavi2017universal,mopuri2018generalizable,thys2019fooling} have been also studied which can change the DNN prediction to a target class for any sample input. 
While Trojan attack is a training-time attack, the universal adversarial attacks do not require training-time access to the model. Further, the trigger in Trojan is known and deliberately injected into the poisoned DNN while these are found through optimization for universal adversarial attacks. While both of these vulnerabilities are a consequence of poor generalization and low resilience of the DNNs, Trojaned DNNs learn to predict a target class for inputs with the trigger. We use attributions to detect these learned triggers. 

The state-of-the-art Trojan insertion 
methods~\cite{edraki2020odyssey, li2020backdoor, schwarzschild2020just, zhang2020backdoor, gu2017badnets,chen2017targeted, gu2017badnets} 
use a minuscule amount of data poisoned with the Trojan trigger pattern (e.g., a local patch, a filter with specific settings).
Alternative methods inject Trojans through transfer
learning~\cite{wang2020backdoor},
retraining a DNN~\cite{liu2017trojaning},
direct manipulation of DNN
weights~\cite{dumford2018backdooring,rakin2020tbt},
or addition of malicious modules~\cite{tang2020embarrassingly}. 
Our  approach 
is independent of the Trojan insertion method (see \autoref{sec:background}).

A number of defense methods have been proposed against Trojan attacks. The first class of defenses are inference-time preprocessing of the samples 
using pre-trained autoencoders~\cite{liu2017neural},
style transfer~\cite{villarreal2020confoc}, 
spatial transformations such as shrinking and flipping~\cite{li2020rethinking}, and the 
superimposition of various image patterns and observation of the conformance of prediction~\cite{gao2019strip}. 
As demonstrated by the US IARPA/NIST-TrojAI datasets, injected triggers can be made robust to many kinds of transformations. 
Our defense approach detects a Trojaned model by analyzing it and not by preprocessing inputs to the model.
Another set of techniques is based on post-processing the trained DNN by retraining it with a set of clean samples
to prune and finetune the DNN~\cite{liu2018fine},
repair of DNN based on the mode connectivity technique~\cite{zhao2020bridging,garipov2018loss}. In contrast, we focus on detecting Trojaned models with a small number of clean samples which are not sufficient for retraining and repair. 
Yet another class of defense methods focus on reverse engineering the Trojan triggers implemented in Neural Cleanse~\cite{wang2019neural} and 
DeepInspect~\cite{chen2019deepinspect}. These trigger generation methods have been shown to produce patterns distinct from the ones used in 
training~\cite{qiao2019defending}. 
GAN based reverse engineering of triggers~\cite{zhu2020gangsweep} have also been proposed. While these approaches do well on stamp-like localized triggers, we consider a large variety of Trojan triggers that include image  filters and hence, cannot be reverse engineered without unreasonable assumptions on prior knowledge of the nature of the trigger. 
We instead adopt an approach to diagnose the DNN for the presence of Trojan triggers. Model diagnosis approaches for Trojan detection include 
the use of universal litmus test~\cite{kolouri2020universal},
differential privacy~\cite{du2019robust},
one-pixel signature~\cite{huang2020one},
and a combination of adversarial attacks and feature inversion~\cite{wang2020practical,zhang2020cassandra}. 
In contrast to these methods, our approach does not rely on susceptibility to adversarial perturbations which can be moderated through adversarial training but instead makes  use of attributions over a few clean samples. We experimentally compare against the state-of-art 
approaches~\cite{wang2019neural,zhang2020cassandra,kolouri2020universal} in Section~\ref{sec:exp}.




\section{Problem Definition}
\label{sec:background}

Given a deep learning model $f$, 
if the Trojan insertion on sample $x$ produces poisoned sample
$x_p$, then the trigger-insertion
relation  $P(x,x_p)$ is said to be true. This relation models multiple poisoned samples that can be created from a single clean sample. 
For any input $x$ and poisoned input $x_p$ with the source class $s$ and the target class $c_t$, $f(x) = s, \forall x \; P(x,x_p) \Rightarrow f(x_p) = c_t $. In contrast to many existing models of triggers as a transformation function, we model it as a relation because trigger transformation need not be a unique function. For example, a polygon trigger can be of different shape, color, size and position, and a filter trigger can be active over a wide set of filter parameters. 
For a trigger to be exploitable in practice, it must be robust to small perturbations. 
We also expect the trigger transformation to satisfy some notion of {\it smallness} in change to avoid detection by online input filtering methods. 
These notions of smallness include $L_l$ norm ($||x_p - x||_l$) distance for $l=1,2,\infty$, and application of Instagram filters such as Gotham and and Nashville ( e.g. US IARPA/NIST-TrojAI dataset). Filters represent changes which might be large in $L_p$ norm space but these perturbations capture physically realizable input variations.

A Trojan detection method is required to identify if a DNN $f$ over prediction classes $C = \{c_1, c_2, \ldots, c_m\}$ contains a Trojan with a perturbation relation $P$, source classes $ C_s \subseteq C$, and the target class $c_t \in C \setminus C_s $, that is, for all inputs $x$ in the input domain $\mathcal D$ of the model, 
 \begin{align*}
\forall x_p \; P(x,x_p) \; \; [ f(x) \in C_s   \Rightarrow f(x_p) = c_t ] 
\end{align*}
Trojaned models have good performance on the inputs without trigger. 
One common approach to inject a Trojan in a DNN is to train it using 
 a mixture of clean and poisoned inputs. This can be achieved by using a modified loss 
 function where $c_x$ denotes the true class of input $x \in \mathcal D$,  $x_p$ is a poisoned input, that is, $P(x,x_p)$ is true, and  $\mathcal{L}$ is the cross-entropy loss function $
\mathcal{L} = \sum_{x}  \mathcal{L}(f(x), c_x) + \sum_{x_p} \mathcal{L}(f(x_p), c_t)$. 
Typically, we do not know the trigger perturbation $P$, the source classes $C_s$ or the target class $c_t$. 

Our attack model has the following characteristics to ensure its relevance to the real-world challenges:
\begin{itemize} [leftmargin=*, noitemsep] 
\item No access to the training data.
\item No knowledge of the trigger perturbation. Our experiments include polygon and filter triggers to demonstrate generalizability of the approach. 
\item No knowledge of the source classes over which the insertion of Trojan trigger would predict the target class. 
\item No knowledge about the target class of the Trojan attack. 
\item No knowledge about the training method used in training the DNN with poisoned data. 
\end{itemize}
We assume the following for our defense approach:
\begin{itemize} [leftmargin=*, noitemsep] 
    \item Assumption 1: We have whitebox access to the DNN - the architecture and the learned weights.
    \item Assumption 2: We have a small set of clean inputs which could be different from the training data.
    \item Assumption 3: The Trojaned model is robust with respect to small changes in the trigger.
\end{itemize}
These characteristics allow us to model adversaries such as outsourced providers of deep learning models using proprietary or privacy-sensitive training data or
proprietary algorithms.
The first two assumptions are reasonable for verifying a DNN acquired from untrusted source. We can expect to have a small set of clean inputs for which we know the true labels and on which the model performs correctly even if it is Trojaned. The third and the final assumption is important for a real-world robust attack so that the attacker can effectively use the trigger to change model's prediction in a noisy environment.

\begin{figure*}[h]
	\begin{center}
        \includegraphics[width=0.8\textwidth]{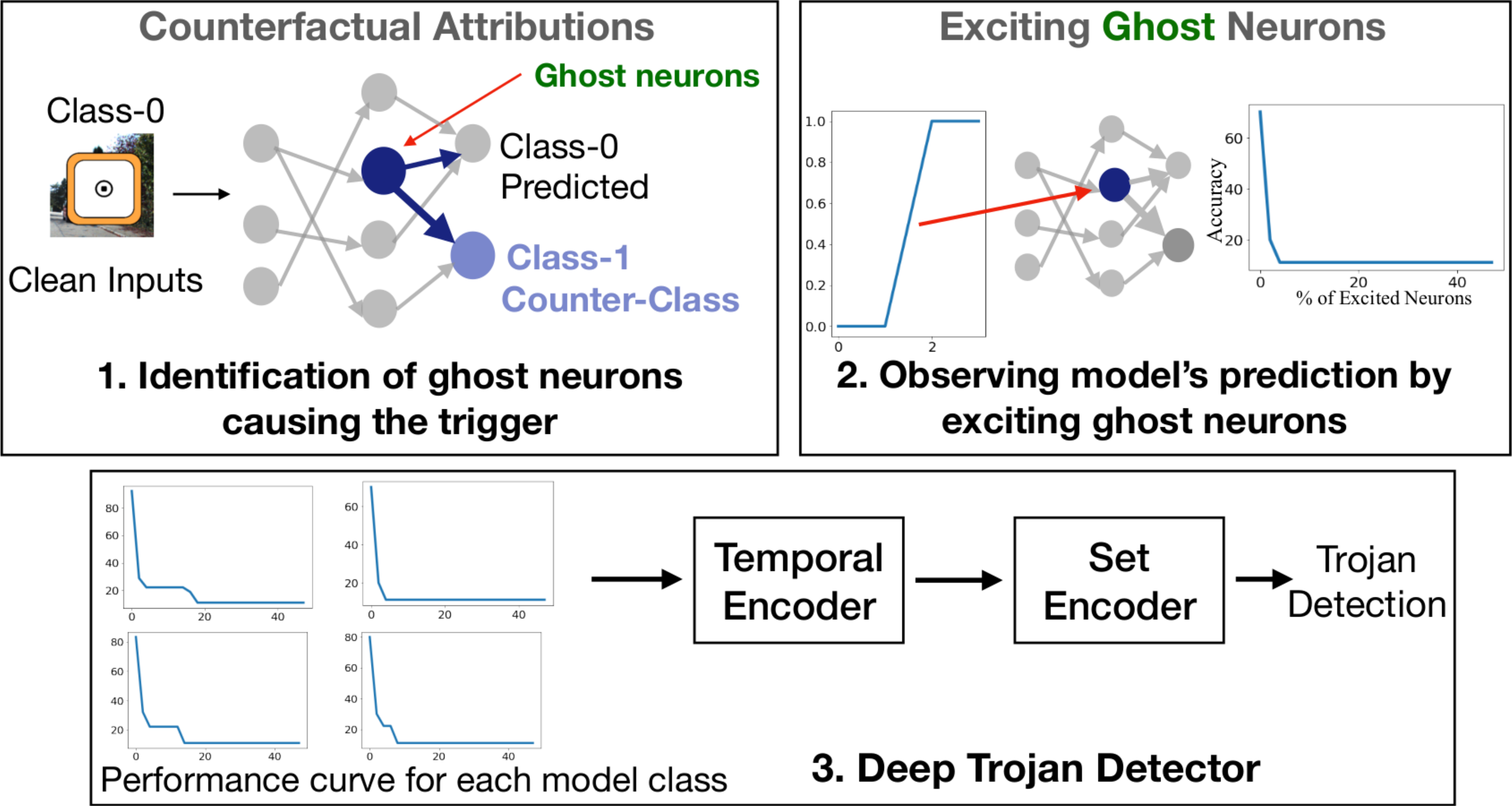}
	\end{center}
	\caption{Showing our approach for Trojan detection. First step identifies ghost neurons, responsible for Trojaned behavior, through counterfactual attributions for each possible target class. It then incrementally excites these ghost neurons and monitor model's accuracy. It finally uses a deep temporal set encoder to input the class-wise performance changes and detect Trojaned DNNs.} 
	\label{image:example_2}
  \end{figure*}
  
\section{Trojaned DNN Detection Approach}
\label{sec:tech}

Our approach is agnostic to the Trojan injection mechanism used by the adversary. 
The first step in our approach is the identification of the suspicious neurons that might cause the Trojaned behavior. 
This is accomplished through 
feature attributions for counterfactual classes other than the predicted output class of the model over the clean samples. 
We call such attribution as \colons{counterfactual attribution} since these measure the contribution of the features in predicting a counter-class that is not the (predicted) highest scoring class \cite{hendricks2018generating, wang2020scout,goyal2019counterfactual}.  
The neurons which have high attribution consistently across the clean samples for the same counterfactual class likely encode the trigger. We refer to these suspect neurons as  
{\it ghost} neurons since they indicate the presence of trigger without it being active in clean samples. These ghost neurons have concise counterfactual attribution as they correspond to the previously reported shortcuts in Trojaned DNNs that allow the trigger to flip the DNN's decision. The second step of our approach excites
these ghost neurons and observes changes in model's accuracy. If these truly encode the trigger and the DNN is Trojaned, the accuracy of the model falls quickly since the trigger gets activated. The fall in accuracy for benign DNNs is more gradual. Finally, we take the class-wise changes in accuracy on excitation of ghost neurons and use a deep set encoder to make the final prediction about the DNN being Trojaned. The deep set encoder ensures that our approach is insensitive to the diversity in the DNNs. We describe each of these steps below.

\subsection*{Attribution-based Ghost Neurons Identification }

\noindent We  use the penultimate layer neurons for input $x$ to a DNN. 
 These neurons are denoted by $z$ and each feature by $z_i$.
Our approach can be applied to any projection of $x$ over a feature space including directly using the pixels. The use of penultimate features has the advantage of using semantically meaningful features. 
We first show that the Assumption 3 on the robustness of trigger implies that the attributions for the features encoding this trigger will be concentrated. This will, in turn, explain the observed quick deterioration in accuracy of the DNN when these features are excited in Trojaned models. 

For robustly trained Trojaned model, we expect the model to produce the correct output on clean data  $f(x) = c_x$ and its perturbations  $f(x+\delta) = c_x$, and produce the target output $c_t$ on poisoned data $f(x_p) = c_t$ and its perturbations $f(x_p + \delta) = c_t$ for the robustness threshold $\delta$. The result is summarized in the theorem below where the attributions of the features are simply the weights $w$ of the last layer $z$ of the DNN. 

\begin{theorem}
The stochastic gradient descent update for robustly training a Trojaned DNN concentrates the   attributions over a small set of features encoding the trigger. 
\end{theorem}

\begin{proof}
The loss function for training a Trojaned model  is  $\mathcal{L} = \sum_{x}  \mathcal{L}(f(x), c_x) + \sum_{x_p} \mathcal{L}(f(x_p), c_t) $. 
Without loss of generality, let us assume that there are only two classes $1$ and $-1$. In order to make the Trojaned DNN robust to $\delta$ perturbations, the loss function is modified to minimize $\mathcal{L}(f(x_p + \delta), c) $ for an input $x_p$.
Typical loss functions such as negative log likelihood or hinge-loss can be written in the form of $g(-c \langle z, w \rangle)$ where $g$ is a non-decreasing function.
Let a subset of the features $s \subseteq z$ correspond to the trigger. 
For each of the features, the expected SGD update for each $w_i$ is $\Delta_i = - \mathop{\mathbb{E}} [ \partial \mathcal{L}(f(x_p + \delta), c) / \partial w_i ] = \mathop{\mathbb{E}} [ g'(\delta_w |w| - c \langle z, w \rangle) (c z_i - sgn(w_i) \delta_w ) ]$ where $\delta_w$ is the worst-case perturbation in $z$ corresponding to the change $\delta$ in $x_p$. We now consider the quantity $\Delta_s = \sum_{i \in s} w_i \Delta_i / \sum_{i \in s}| w_i | $ which has a natural interpretation as change in concentration of the attributions. The high positive value of $sgn(w_i) \Delta_i$ means expansion while high negative value means shrinkage. $\Delta_s$ models the weighted expansion or shrinkage across attributions. Further, 
$\Delta_s \leq \mathop{\mathbb{E}} [ g'(\delta_w |w| - c \langle z, w \rangle) (\gamma_s -  \delta_w ) ] $ where $\gamma_s = c \sum_{i\in s} w_i z_i / \sum_{i\in s} |w_i| $ is the output $c$ aligned weighted strength of the features. 
If $\gamma_s$ is not aligned, $\Delta_s$ pushes the attributions to $0$ and  
if $\gamma_s$ is aligned but not concentrated enough to be larger than $\delta_w$, then 
$\Delta_s$ still pushes the attributions to $0$ making the attributions more concentrated. Thus, as we increase the robustness $\delta_w$ of the trigger, the features corresponding to the Trojan become further concentrated.
\end{proof}
\noindent We make the following observations that motivate our technical approach. The first observation is used to develop our Trojaned model detection approach and follows from the concentration of attributions. The second central observation is the manifestation of these features as having high attribution even on clean data when looking at the counterfactual class which matches the trigger's target class. This motivates our counterfactual analysis and enables us to find Trojans without the need for poisoned samples. 

\paragraph{Observation 1}: The robust triggers in a Trojaned model are encoded using a few features in the penultimate layer of a Trojaned model. 

\paragraph{Observation 2}: When examining the counterfactual attribution over the features in the penultimate layer on decisions on clean samples , if the counterfactual class is the target class in a Trojaned model, these ghost neurons encoding the trigger exhibit high attribution. \\

\subsection*{Shortcut Pathway in Feature Space}

\noindent We draw a connection between the approach proposed in this paper and 
a common hypothesis shared in literature for Trojan detection \cite{wang2019neural,zhang2020cassandra}. This hypothesis 
states that a \textbf{shortcut pathway} is present inside a Trojaned model and enables the model to predict target class on poisoned input without affecting its performance on clean samples. Using the above two observations, 
 we refine this hypothesis by identifying that such shortcuts are in the form of a collection of a small set of features that activate the trigger behavior. These shortcuts in the model persist even for clean samples and counterfactual analysis can be used to get attributions over these trigger features. This is in contrast to indirect detection of these shortcuts using adversarial attacks because it is not necessary that adversarial examples exploit this shared common shortcut in the model instead of identifying sample-specific perturbations in the case of individual attacks and other perturbations (not the actual injected trigger) in case of universal attacks \cite{zhang2020cassandra, kolouri2020universal}. We  compute the attribution of the features for different class outputs enabling us to directly detect these shortcut pathways and locate the ghost neurons. We use counterfactual explanations 
to rank neurons on the degree of their contribution to a target class.  
More generally, the shortcut pathway and the ghost neurons can occur across multiple layers \cite{olah2020zoom} but our experiments demonstrate that analyzing the penultimate layer is sufficient for detecting Trojaned models. 
 We identify the ghost neurons by using attributions over input features.
 We denote the attribution for a neuron $z$ (drop subscript $i$) for class $c_k$ as $\alpha^z_k(\mathcal{\mathcal{X}}) = \frac{1}{|\mathcal{X}|}\sum_{x_i \in \mathcal{X}} Attr(f(x_i), c_k, z)$, where $Attr$ is the attribution function and $\mathcal{X}$ is a dataset containing some samples. If we had access to the poisoned inputs, we could have used them to compute attributions and directly identify ghost neurons.  

\begin{figure}[!h]
	\begin{center}
        \includegraphics[width=0.4\textwidth]{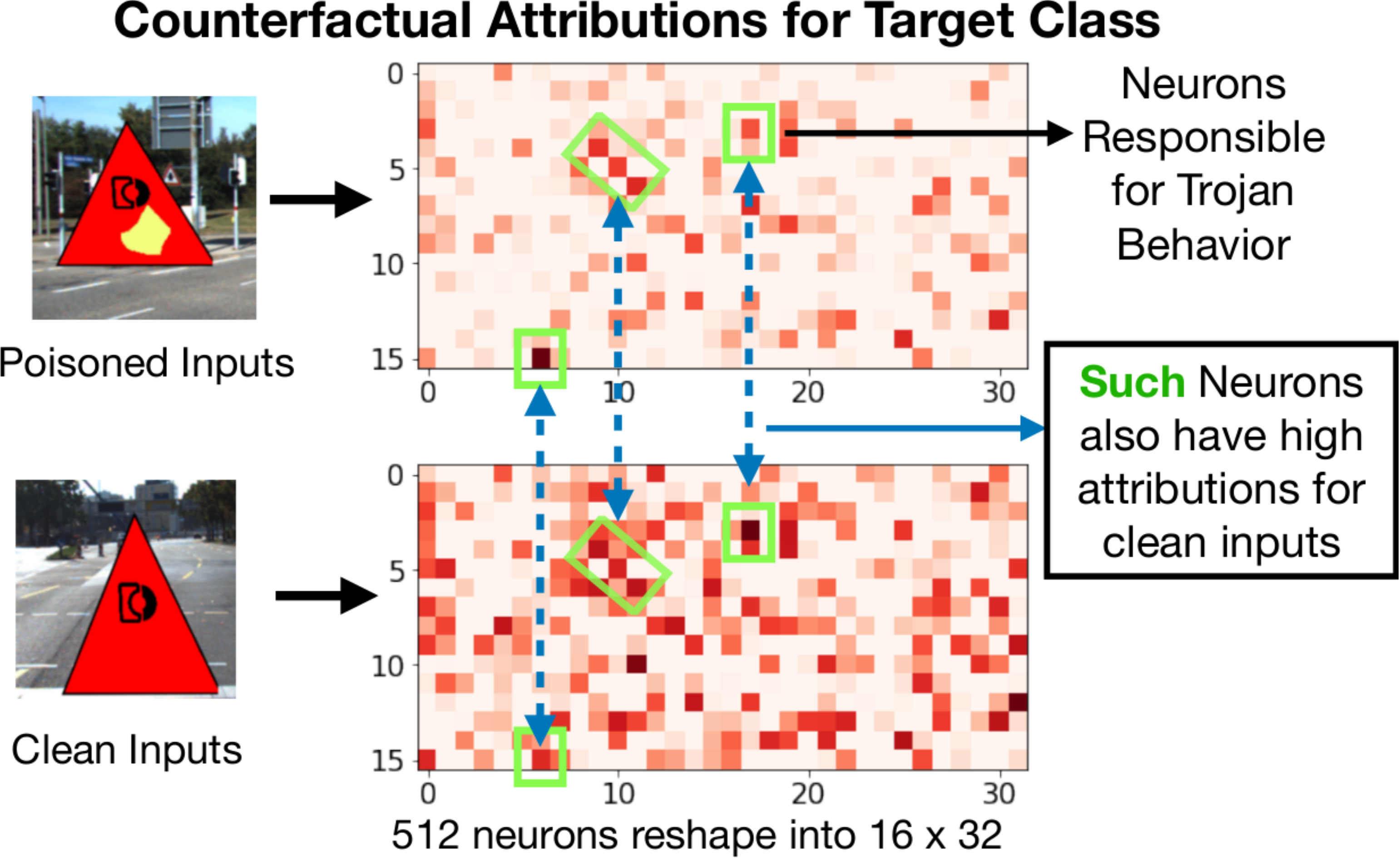}
	\end{center}
	\caption{Showing neuron attributions for predicting the target class across poisoned and clean inputs for a Trojaned model. We observe that neurons causing the Trojan behavior (ghost neurons) also have high counterfactual attributions (darker red implies higher) across clean inputs.}
	\label{image:attr}
  \end{figure}
  
\subsection*{Counterfactual Analysis: Exciting Ghost Neurons}
\label{sec:ghost-neurons}

\noindent However, we do not have access to poisoned inputs in real-world settings. We thus modify our approach to rank neurons by instead computing counterfactual attributions across clean images for predicting target class $c_k$ \ie{} $\alpha^z_k(\mathcal{X}^c)$, where $\mathcal{X}^c$ is the dataset of clean inputs. 
A key reason that we are able to use the clean images as proxies for the poisoned images is because the ghost neurons are poly-semantic in nature,  
where they fire for patterns corresponding to both actual class(es) and the trigger pattern. 
This allows us to estimate the neuron rankings identified with poisoned images within some error margin, 
which are then used for our next analysis. 
\autoref{image:attr} shows attributions $\alpha^z_t$ for the target class across poisoned and clean inputs for a Trojaned model. We observe that neurons with high attributions for poisoned inputs also have high (counterfactual) attributions for clean inputs, which allows us to identify them.   

\begin{figure}[h]
	\begin{center}
        \includegraphics[width=\columnwidth]{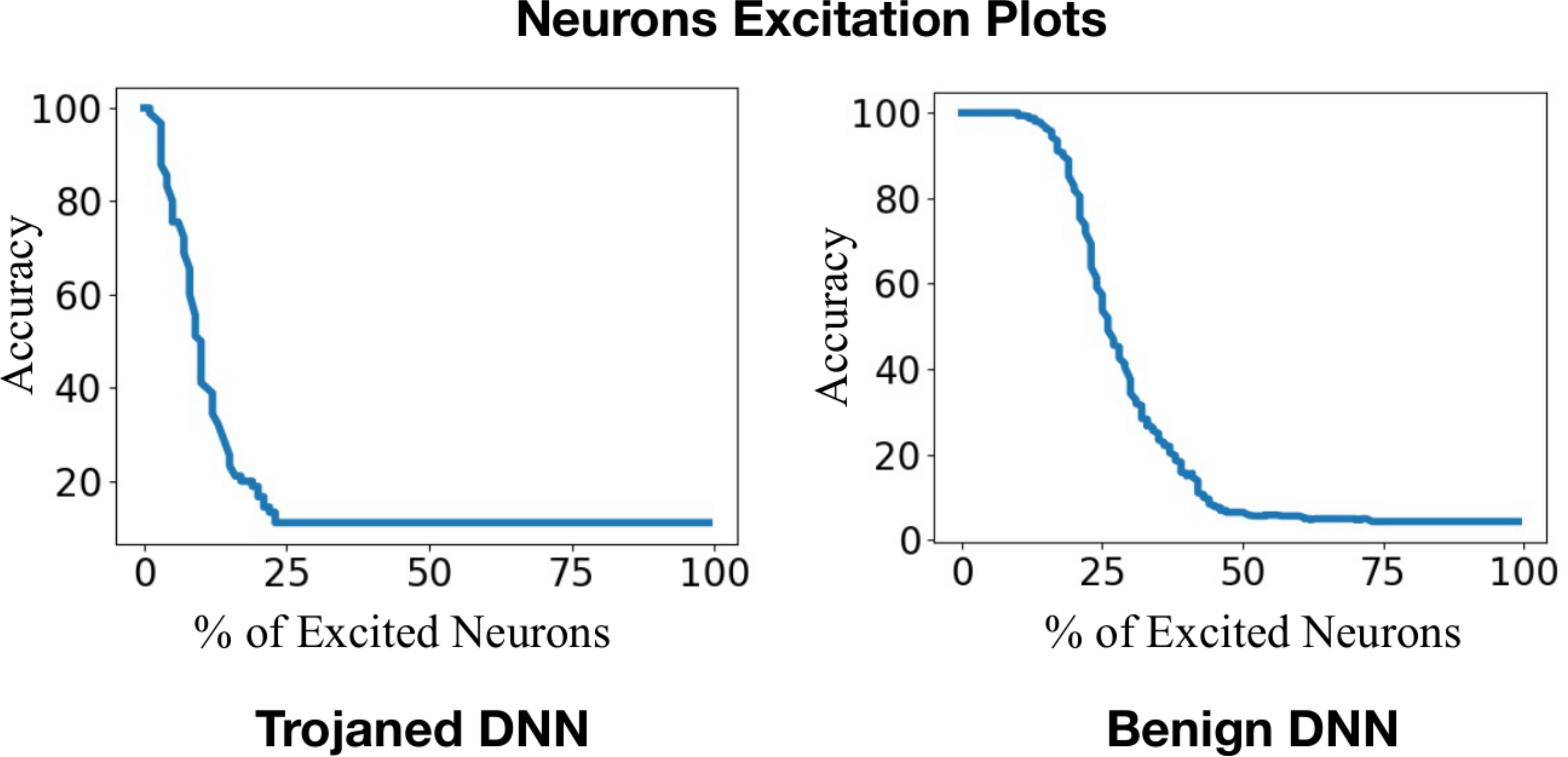}
	\end{center}
	\caption{The ghost neurons identified using the counterfactual analysis are excited incrementally to monitor accuracy. Figure highlights that the trigger, if present, is localized in a few neurons in a Trojaned DNN as compared to a benign DNN.}
	\label{image:ablation}
  \end{figure}

For the next step, we use the Observation 1 about sparseness of the ghost neurons. 
This is a critical property that separates ghost neurons from normal neurons and thus benign models from Trojaned models.  
We exploit this observation by computing the DNN's performance on the clean samples by gradually exciting neurons (with twice the maximum activation across all the neurons in the penultimate layer) based on their ranks.  
If the model is Trojaned, a sharp drop in performance is observed on excitation of a small number of neurons.   
For example, \autoref{image:ablation} shows the model's performance versus percentage of excited neurons for a benign model and for a Trojaned model. It is evident that the fall in performance is sharp for the Trojaned model.

\paragraph{Addressing Model Diversity Using a Deep Temporal Set Encoder}
\noindent Since the target class is unknown, the above approach is repeated for the DNN classes. We can then pick the possible target class based on the steepest descent in the performance curve and extract relevant features such as the rate of fall in performance. However, this does not generalize to DNNs with different number of classes, architectures, training regimes, \etc 
We address this challenge by proposing a deep temporal set encoder based Trojan detector that contains the right inductive biases to encode the class-wise performance curves.
Our model first encodes the performance curves as time-series by using a temporal encoder, that is either a 1D-CNN or a Transformer model with positional encoding. 
The temporal encoder produces a tensor containing features for all the classes. 
Next, we use a set encoder that is
invariant to the number and ordering of classes since different DNN can have different number of classes.
We achieve this by using a permutation invariant encoder that treats the class-level outputs as sets of features and 
pools their outputs using max-pooling \cite{zaheer2017deep, qi2017pointnet}. We pass the pooled output
through a linear layer for prediction. We then train our model on a dataset containing both benign and Trojaned DNNs using cross-entropy loss.

\section{Experiments}
\label{sec:exp}

We evaluate our approach  on four datasets containing Trojaned DNNs trained for image classification. 
Our approach is implemented in the tool
{\it TrinityAI} that aims at addressing the three interrelated challenges of improving Trust, Resilience and Interpretability of AI models. 
We first describe the datasets and the evaluation metrics.
We then discuss qualitative results that also includes comparison with state-of-the-art (SOTA) methods. We finally study the impact of factors that can be varied by an adversary for training Trojaned DNNs on performance.

\subsection{Datasets and Evaluation Metrics}
\label{sec:dataset}

\paragraph{Triggered-MNIST:} We use the code provided by NIST\footnote{\url{https://github.com/trojai/trojai}} to generate 810 DNNs ($50\%$ are Trojaned) 
trained to classify MNIST digits. The DNNs are selected from among three architectures. The Trojaned models are trained with images poisoned with two types of trigger patterns that are randomly inserted into the image. The attack is designed to misclassify some source classes to a target class. Other factors such as the target class and poisoning rate are selected randomly (details in \red{Section 2} in Supplementary Material).

\paragraph{TrojAI-Round1, Round2, Round3:} These datasets are made publicly available by US IARPA/NIST\footnote{\url{https://pages.nist.gov/trojai/docs/data.html}} and contains models trained for traffic sign classification ($50\%$ are Trojaned). The models are trained on synthetically generated image-data of artificial traffic signs superimposed on road background scenes. Trojan detection is harder on Round2/Round3 in comparison to Round1 due to larger variations in factors used for training Trojaned models such as (1) number of classes-- $5$ in Round1 versus $5-25$ in Round2/Round3, (2) trigger types-- polygon triggers in Round1 versus polygon and Instagram filter based triggers in Round2/Round3, (3) the number of source classes-- all classes are poisoned in Round1 versus 1, 2, or all classes in Round2/Round3, and (4) the number of model architectures--  $3$ in Round1 versus $23$ in Round2/Round3. 
Compared to Round2 models in Round3 are adversarially trained using two methods--Projected Gradient Descent and Fast is Better than Free \cite{wong2020fast}. The polygon trigger is generated randomly with variations in shape, size, and color. The filter based trigger is generated by randomly choosing from five distinct filters. Trojan detection is much harder on the TrojAI datasets as compared to the Triggered-MNIST dataset due to the use of deeper DNNs and larger variations in appearances of foreground/background objects, trigger patterns \etc. Round1, Round2, and Round3 have $1000$, $1104$, and $1008$ models respectively.

\begin{table*}[!htbp]
  \centering
  \begin{tabular}{c|c|c|c|c}
      \hline
      Model & Triggered- & TrojAI- & TrojAI-  & TrojAI-    \\
       & MNIST & Round1 & Round2 & Round3 \\  \hline  
      Cassandra \cite{zhang2020cassandra} & $\mathbf{0.97\pm0.010}$ & $0.88\pm0.006$ & $0.59\pm0.096$ & $0.71\pm0.026$ \\ 
       Neural Cleanse \cite{wang2019neural} & $0.70\pm0.045$ & $0.50\pm0.030$ & $0.63\pm0.043$ & $0.61\pm0.064$ \\ 
       ULP \cite{kolouri2020universal} & $0.54\pm0.051$ & $0.55\pm0.058$ & $-$ & $-$ \\ \hline
       \trinity-Conv-IG & $0.89\pm0.024$ & $0.87\pm0.020$ & $0.73\pm0.014$ & $0.71\pm0.038$ \\
       \trinity-Tx-IG & $0.95\pm0.022$ & $0.89\pm0.029$ & $0.75\pm0.033$ & $\mathbf{0.72\pm0.038}$  \\
       \trinity-Conv-GradxAct & $0.87\pm0.030$ & $0.88\pm0.027$ & $0.74\pm0.030$ & $0.67\pm0.036$ \\ 
       \trinity-GradxAct & $0.96\pm0.014$ & $\mathbf{0.90\pm0.027}$ & $\mathbf{0.76\pm0.027}$ & $0.66\pm0.029$ \\          
  \end{tabular}
  \vspace{0.5em}
  \caption{Comparison of our model (\trinity) with three SOTA method on four datasets containing Trojaned DNNs. \colons{Conv} and \colons{Tx} refers to CNN and Transformer based temporal encoder. \colons{IG} and \colons{GradxAct} refers to attribution methods Integrated Gradients \cite{sundararajan2017axiomatic} and gradient times activation. Performance is reported using mean AUC with standard deviation across $5$ random splits. 
  }
  \label{table:quant}
\end{table*}

\paragraph{Metrics:} We report mean and standard deviation of area under the ROC curve (AUC) on $5$ randomly selected splits with $80\%$ of the models for training, $10\%$ for validation, and $10\%$ for testing. 
We provide implementation details for different methods in the Supplementary (\red{Section 1}).


\subsection{Quantitative Results}

\autoref{table:quant} shows the performance of our approach (\trinity) along with three SOTA methods on four datasets. 
It also shows the performance of \trinity\; with two different temporal encoders and attribution methods. \trinity-Tx-IG and \trinity-Conv-IG use Integrated Gradients as the attribution method, and the temporal encoder is based on Transformer and Convolution, respectively. Similarly, \trinity-Tx-GradxAct and \trinity-Conv-GradxAct use different temporal encoders, and the attribution method is GradientxActivation.

\paragraph{Ablation:} We observe a drop in performance when going from datasets with smaller variations in model architectures and trigger types (Triggered-MNIST, TrojAI-Round1) to those with larger variations (TrojAI-Round2/Round3).  
For example, \trinity-Tx-IG achieves an AUC of $0.95$, $0.89$, $0.75$, and $0.72$ on Triggered-MNIST, TrojAI-Round1, Round2, and Round3 respectively. This drop is expected as Trojan detection becomes more challenging on datasets with more variations among DNNs.
We observe that the temporal encoder based on Transformer performs better than CNN in most cases. 
The gains are higher for Triggered-MNIST ($0.95$ of \trinity-Tx-IG versus $0.89$ of \trinity-Conv-IG) as compared to TrojAI datasets ($0.75$ versus $0.73$ on Round2). This is the case since TrojAI datasets contain DNN with larger variations in model architecture, trigger pattern \etc{}, which reduces the effective data-points across each factor leading to saturation of the Transformer. It might be helpful to use more data or smart pre-training strategies as used in BERT \cite{vaswani2017attention, devlin2018bert}. 
We also experimented with two different attribution approaches-- Integrated Gradients and GradxAct which multiplies gradients (with respect to input) with activations. 
We see minor improvements with GradxAct method on all datasets except TrojAI-Round3 \eg{} AUC changes from $0.89$ (\trinity-Tx-IG) to $0.90$ on Round1 and $0.72$ to $0.66$ on Round3  
We believe the performance falls on Round3 due to the use adversarial training which reduces the effectiveness of simpler attribution methods (such as GradxAct) in being able to recover the ghost neurons. We plan to investigate other attribution methods such as Shapley values in the future.  
Overall these results highlight the efficacy of our approach in identifying the ghost neurons using clean inputs and using them Trojan detection.

\paragraph{Comparison with SOTA}:We compare our approach with SOTA methods that diagnose DNNs through sensitivity to universal attacks (Cassandra \cite{zhang2020cassandra}), generated noise patterns (ULP \cite{kolouri2020universal}), and reverse engineered triggers (Neural Cleanse \cite{wang2019neural}). We noted earlier that these methods make strong assumptions regarding the nature of the attack or the ability of adversarial inputs to activate the shortcut pathway, which limits their generalizability to real-world datasets. On Triggered-MNIST, where DNNs belong to three  model architectures and are poisoned with only polygon triggers, Cassandra achieves an AUC of $0.97$ as compared to $0.95$ of \trinity-Tx-IG. However, the performance drops drastically compared to \trinity-Tx-IG as we move from Round1 ($0.88$ versus $0.89$) to Round2 ($0.59$ versus $0.75$). This happens since the universal adversarial attack used in Cassandra is unable to localize the shortcut pathway in Round2, where DNNs belong to $23$ model architectures and are poisoned with two different trigger types.   
Cassandra performs similar to our model on Round3 where the DNNs are adversarially trained, which probably increases the chances of universal attacks to trigger the shortcut pathway. Our model does not require such adversarial training.  
Our model also outperforms both ULP and Neural Cleanse by a large margin \eg{} AUC is $0.55$ for ULP, $0.50$ for Neural Cleanse, and $0.89$ for \trinity-Tx-IG on Round1. Both of these methods make further assumptions on the nature of the attacks and thus fail to generalize to real-world datasets. We observe that Neural Cleanse performs better on Round2/Round3 compared to Round1. We believe this happens since Neural Cleanse uses anomaly detection on the $L1$ norm which works better for models with a larger number of classes.    
We would also like to note that compared to Neural Cleanse and ULP, that required making changes to the attack parameters for each dataset, \trinity\; uses the same feature extraction for all datasets. This highlights the strong generalization capability of our approach.

\subsection{Impact of Factors Controlled by Adversary}
\label{sec:analysis}

\begin{table}[t!]
  \centering
  \scalebox{0.84}{
  \begin{tabular}{|c|c|}
  \hline
    Trigger & AUC \\
    Type & \\ \hline
    Polygon & $0.83\pm0.033$  \\ \hline
    Filter & $0.65\pm0.043$ \\ \hline 
  \end{tabular} 
  } 
  \scalebox{0.84}{
  \begin{tabular}{|c|c|}
    \hline
    No. of Source & AUC \\ 
    Classes & \\ \hline
    1 & $0.55\pm0.063$  \\ \hline
    2 & $0.63\pm0.038$ \\  \hline
    all & $0.95\pm0.022$ \\ \hline 
  \end{tabular}
  }

  \scalebox{0.84}{
  \begin{tabular}{|c|c|}
    \hline
    Model & AUC \\ 
    arch & \\ \hline
    ResNet50 & $0.92\pm0.049$  \\ \hline
    InceptionV3 & $0.95\pm0.033$ \\ \hline
    DenseNet121 & $0.86\pm0.064$ \\ \hline
    all & $0.88\pm0.076$ \\ \hline 
  \end{tabular}} 
  \scalebox{0.84}{
  \begin{tabular}{|c|c|}
    \hline
    Poisoning Rate & AUC \\ \hline
    $>=0.29$ & $0.78\pm0.045$  \\ \hline
    $<0.29$ & $0.62\pm0.058$ \\ \hline
    
  \end{tabular}}
  \caption{Impact of factors, that can be varied by an adversary for training Trojaned DNNs, on our model's performance.}
  \label{table:ablation}
  
\end{table}

We wish to investigate the impact of factors, that can be varied by an adversary for training Trojaned DNNs, on our model's performance.  
We focus on four key factors-- trigger type, number of source classes, model architecture and poisoning rate.  
For each factor we create equal-sized subsets whose DNNs assume a fixed value for that factor and then evaluate against \trinity-Conv-IG in \autoref{table:ablation}.

\paragraph{Trigger-Type:} We create two subsets from Round2 containing Trojaned models with either polygon or filters based triggers. 
AUC with polygon and filter based triggers is $0.83$ and $0.65$ respectively. The performance for polygon triggers is higher since  
it is easier to identify the ghost neurons for triggers that are well localized in the image. 
On the other hand, filter based triggers are harder to localize as they are distributed across the entire image. This rationale also explains the drop in performance for all methods between Round1 and Round2, as the later contains DNNs poisoned with filter based triggers. 

\paragraph{Number of source classes:} We create three equal-sized subsets from Round2, each containing Trojaned models with $1$, $2$ and all classes as source classes (classes whose inputs are misclassified when poisoned). The AUC consistency improves across $1$, $2$ and all classes ($0.55$, $0.63$, and $0.95$ respectively) as      
the shortcut pathway becomes better visible as more classes are poisoned, making it easier to identify the ghost neurons from the clean data.

\paragraph{Model architectures:} We create four equal-sized subsets from Round1 containing all benign models and $250$ randomly sampled Trojaned models from ResNet, InceptionV3, DenseNet121, and all three architectures. The AUC is $0.92$, $0.95$, $0.86$, and $0.88$ for for ResNet50, InceptionV3, DenseNet121, and all architecture respectively. This shows that the performance of the model depends on the model architecture. 
This is probably because the architectural differences affect the way information is encoded by neurons \cite{bau2017network}. 
For example the performance is higher for models 
with intra-layer skip-connections such as ResNet and InceptionV3. However, the performance is lower for complex models (DensNet121) that contains inter-layer skip-connections due the difficulty in localizing the ghost neurons. 
Also, it is harder to generalize across multiple models resulting in the lowest performance for \colons{all} architecture.

\paragraph{Poisoning Rate:} We create two equal-sized subsets from Round2 with poisoning rate (\% of poisoned data used during training) $>=0.29$ and $<0.29$ respectively. The AUC is higher for poisoning rate $\geq 0.29$ ($0.78$ vs $0.62$) since the shortcut pathway is more visible in this case.

\begin{figure}[tbp!]
	\begin{center}
        \includegraphics[width=1\columnwidth]{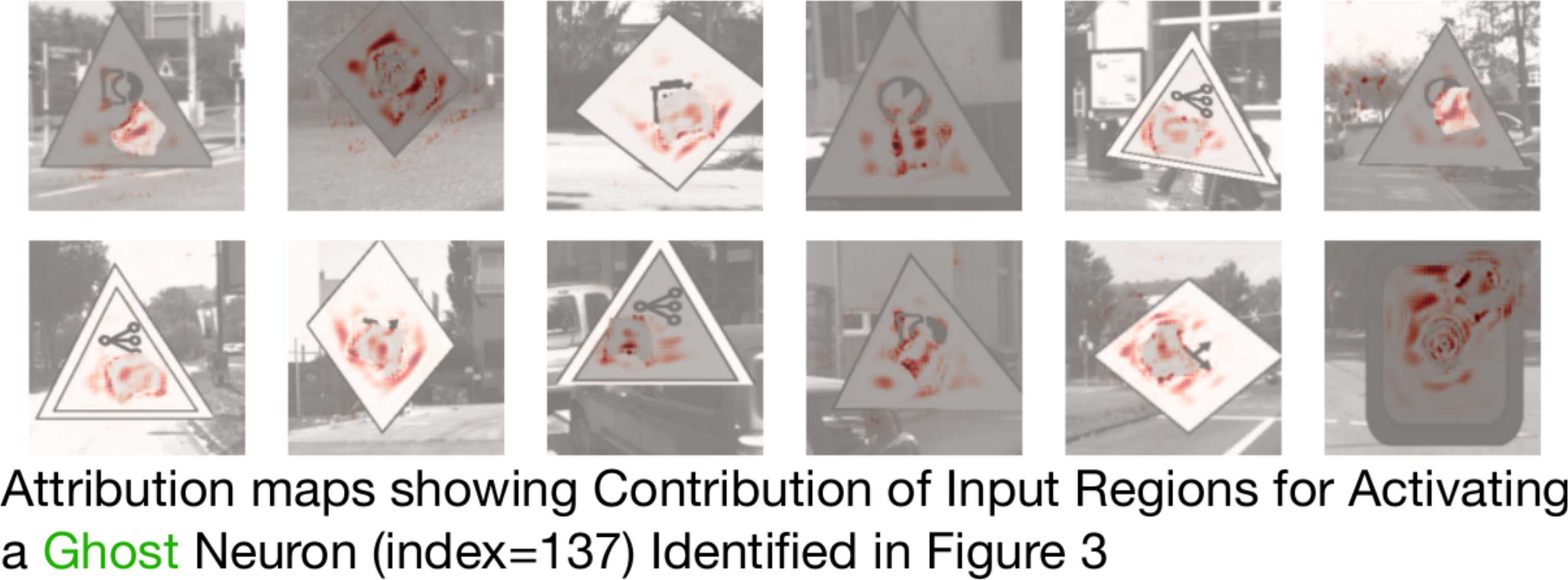}
	\end{center}
	\caption {Heatmaps reveal the poly-semantic nature of a ghost neuron that is responsible for the trigger behavior and fires for patterns from both the actual class(es) and the trigger.}
	\label{image:neuron_attr}
  \end{figure}

\autoref{image:neuron_attr} shows attribution heatmaps that reveal the contribution of image parts for activating one of the ghost neurons identified in \autoref{image:attr}. We observe that the high attribution parts include both the actual trigger and other parts of the images. This validates our characterization of the ghost neurons as being poly-semantic, which enables us to identify them using counterfactual attribution on clean inputs.




 
\section{Conclusion}
\label{sec:conclusion}

We focused on predicted Trojaned DNNs through counterfactual attributions over clean inputs. This is a challenging problem since Trojaned DNNs behave similar to benign DNNs except producing incorrect outputs for inputs poisoned with a pre-determined trigger.    
We based our approach on the idea that  
the trigger behavior is localized on a few ghost neurons that are poly-semantic in nature and fire for both input classes and trigger pattern.  
We provided a mathematical basis for this idea and then proposed an approach that uses counterfactual attributions over clean inputs to localize these ghost neurons. We then excite these neurons to observe changes in model's accuracy. We use a deep temporal set encoder to input these class-wise performance curves and train a Trojan Detection network. We evaluated our approach on a set of  challenging benchmarks with large diversity in model architectures, number of classes, trigger pattern etc. Our results show that the proposed approach is able to consistently improve upon state-of-the-art methods especially for challenging datasets. As part of our future work, we plan to better understand the effect of different training strategies and model architectures on the  counterfactual attributions for Trojan detection.

\section*{Acknowledgement}
The authors acknowledge support from  
IARPA TrojAI under 
contract W911NF-20-C-0038. The views, opinions and/or findings expressed are those of the author(s) and should not be interpreted as representing the official views or policies of the Department of Defense or the U.S. Government.

{\small
\bibliographystyle{ieee_fullname}
\bibliography{egbib}

\begin{thebibliography}{10}\itemsep=-1pt

\bibitem{adebayo2018sanity}
Julius Adebayo, Justin Gilmer, Michael Muelly, Ian Goodfellow, Moritz Hardt,
  and Been Kim.
\newblock Sanity checks for saliency maps.
\newblock In {\em NIPS}, pages 9525--9536, 2018.

\bibitem{ancona2017towards}
Marco Ancona, Enea Ceolini, Cengiz {\"O}ztireli, and Markus Gross.
\newblock Towards better understanding of gradient-based attribution methods
  for deep neural networks.
\newblock {\em arXiv preprint arXiv:1711.06104}, 2017.

\bibitem{bahdanau2014neural}
Dzmitry Bahdanau, Kyunghyun Cho, and Yoshua Bengio.
\newblock Neural machine translation by jointly learning to align and
  translate.
\newblock {\em arXiv preprint arXiv:1409.0473}, 2014.

\bibitem{bau2017network}
David Bau, Bolei Zhou, Aditya Khosla, Aude Oliva, and Antonio Torralba.
\newblock Network dissection: Quantifying interpretability of deep visual
  representations.
\newblock In {\em Conference on computer vision and pattern recognition}, pages
  6541--6549, 2017.

\bibitem{carlini2019evaluating}
Nicholas Carlini, Anish Athalye, Nicolas Papernot, Wieland Brendel, Jonas
  Rauber, Dimitris Tsipras, Ian Goodfellow, Aleksander Madry, and Alexey
  Kurakin.
\newblock On evaluating adversarial robustness.
\newblock {\em arXiv preprint arXiv:1902.06705}, 2019.

\bibitem{chalasani2018adversarial}
Prasad Chalasani, Somesh Jha, Aravind Sadagopan, and Xi Wu.
\newblock Adversarial learning and explainability in structured datasets.
\newblock {\em arXiv preprint arXiv:1810.06583}, 2018.

\bibitem{chen2019deepinspect}
Huili Chen, Cheng Fu, Jishen Zhao, and Farinaz Koushanfar.
\newblock Deepinspect: A black-box trojan detection and mitigation framework
  for deep neural networks.
\newblock In {\em International joint conferences on artificial intelligence},
  pages 4658--4664, 2019.

\bibitem{chen2017targeted}
Xinyun Chen, Chang Liu, Bo Li, Kimberly Lu, and Dawn Song.
\newblock Targeted backdoor attacks on deep learning systems using data
  poisoning.
\newblock {\em arXiv preprint arXiv:1712.05526}, 2017.

\bibitem{devlin2018bert}
Jacob Devlin, Ming-Wei Chang, Kenton Lee, and Kristina Toutanova.
\newblock Bert: Pre-training of deep bidirectional transformers for language
  understanding.
\newblock {\em arXiv preprint arXiv:1810.04805}, 2018.

\bibitem{du2019robust}
Min Du, Ruoxi Jia, and Dawn Song.
\newblock Robust anomaly detection and backdoor attack detection via
  differential privacy.
\newblock {\em arXiv preprint arXiv:1911.07116}, 2019.

\bibitem{dumford2018backdooring}
Jacob Dumford and Walter Scheirer.
\newblock Backdooring convolutional neural networks via targeted weight
  perturbations.
\newblock {\em arXiv preprint arXiv:1812.03128}, 2018.

\bibitem{edraki2020odyssey}
Marzieh Edraki, Nazmul Karim, Nazanin Rahnavard, Ajmal Mian, and Mubarak Shah.
\newblock Odyssey: Creation, analysis and detection of trojan models.
\newblock {\em arXiv preprint arXiv:2007.08142}, 2020.

\bibitem{gao2019strip}
Yansong Gao, Change Xu, Derui Wang, Shiping Chen, Damith~C Ranasinghe, and
  Surya Nepal.
\newblock Strip: A defence against trojan attacks on deep neural networks.
\newblock In {\em Computer security applications conference}, pages 113--125,
  2019.

\bibitem{garipov2018loss}
Timur Garipov, Pavel Izmailov, Dmitrii Podoprikhin, Dmitry~P Vetrov, and
  Andrew~G Wilson.
\newblock Loss surfaces, mode connectivity, and fast ensembling of dnns.
\newblock In {\em Neural information processing systems}, pages 8789--8798,
  2018.

\bibitem{goyal2019counterfactual}
Yash Goyal, Ziyan Wu, Jan Ernst, Dhruv Batra, Devi Parikh, and Stefan Lee.
\newblock Counterfactual visual explanations.
\newblock {\em arXiv preprint arXiv:1904.07451}, 2019.

\bibitem{graves2013speech}
Alex Graves, Abdel-rahman Mohamed, and Geoffrey Hinton.
\newblock Speech recognition with deep recurrent neural networks.
\newblock In {\em International conference on acoustics, speech and signal
  processing}, pages 6645--6649, 2013.

\bibitem{gu2017badnets}
Tianyu Gu, Brendan Dolan-Gavitt, and Siddharth Garg.
\newblock Badnets: Identifying vulnerabilities in the machine learning model
  supply chain.
\newblock {\em arXiv preprint arXiv:1708.06733}, 2017.

\bibitem{hao2020adversarial}
Han Xu Yao~Ma Hao-Chen, Liu~Debayan Deb, Hui Liu Ji-Liang~Tang Anil, and K
  Jain.
\newblock Adversarial attacks and defenses in images, graphs and text: A
  review.
\newblock {\em International journal of automation and computing},
  17(2):151--178, 2020.

\bibitem{hendricks2018generating}
Lisa~Anne Hendricks, Ronghang Hu, Trevor Darrell, and Zeynep Akata.
\newblock Generating counterfactual explanations with natural language.
\newblock {\em arXiv preprint arXiv:1806.09809}, 2018.

\bibitem{howard2017mobilenets}
Andrew~G Howard, Menglong Zhu, Bo Chen, Dmitry Kalenichenko, Weijun Wang,
  Tobias Weyand, Marco Andreetto, and Hartwig Adam.
\newblock Mobilenets: Efficient convolutional neural networks for mobile vision
  applications.
\newblock {\em arXiv preprint arXiv:1704.04861}, 2017.

\bibitem{huang2020one}
Shanjiaoyang Huang, Weiqi Peng, Zhiwei Jia, and Zhuowen Tu.
\newblock One-pixel signature: Characterizing cnn models for backdoor
  detection.
\newblock {\em arXiv preprint arXiv:2008.07711}, 2020.

\bibitem{jha2019safeml}
Susmit Jha, Sunny Raj, Steven Fernandes, Sumit~Kumar Jha, Somesh Jha, Jalaian
  Brian, Gunjan Verma, and Ananthram Swami.
\newblock Attribution-driven causal analysis for detection of adversarial
  examples.
\newblock {\em Safe Machine Learning workshop at ICLR}, 2019.

\bibitem{jha2019attribution}
Susmit Jha, Sunny Raj, Steven Fernandes, Sumit~K Jha, Somesh Jha, Brian
  Jalaian, Gunjan Verma, and Ananthram Swami.
\newblock Attribution-based confidence metric for deep neural networks.
\newblock In {\em Neural information processing systems}, pages 11826--11837,
  2019.

\bibitem{jha2017nfm}
Susmit Jha, Vasumathi Raman, Alessandro Pinto, Tuhin Sahai, and Michael
  Francis.
\newblock On learning sparse {B}oolean formulae for explaining {AI} decisions.
\newblock In {\em NASA Formal methods symposium}, pages 99--114. Springer,
  2017.

\bibitem{jha2019jar}
Susmit Jha, Tuhin Sahai, Vasumathi Raman, Alessandro Pinto, and Michael
  Francis.
\newblock Explaining {AI} decisions using efficient methods for learning sparse
  {B}oolean formulae.
\newblock {\em Journal of automated reasoning}, 63(4):1055--1075, 2019.

\bibitem{kilbertus2018generalization}
Niki Kilbertus, Giambattista Parascandolo, and Bernhard Sch{\"o}lkopf.
\newblock Generalization in anti-causal learning.
\newblock {\em arXiv preprint arXiv:1812.00524}, 2018.

\bibitem{kokhlikyan2020captum}
Narine Kokhlikyan, Vivek Miglani, Miguel Martin, Edward Wang, Bilal Alsallakh,
  Jonathan Reynolds, Alexander Melnikov, Natalia Kliushkina, Carlos Araya, Siqi
  Yan, et~al.
\newblock Captum: A unified and generic model interpretability library for
  pytorch.
\newblock {\em arXiv preprint arXiv:2009.07896}, 2020.

\bibitem{kolouri2020universal}
Soheil Kolouri, Aniruddha Saha, Hamed Pirsiavash, and Heiko Hoffmann.
\newblock Universal litmus patterns: Revealing backdoor attacks in cnns.
\newblock In {\em Conference on computer vision and pattern recognition}, pages
  301--310, 2020.

\bibitem{krizhevsky2012imagenet}
Alex Krizhevsky, Ilya Sutskever, and Geoffrey~E Hinton.
\newblock Imagenet classification with deep convolutional neural networks.
\newblock In {\em Neural information processing systems}, pages 1097--1105,
  2012.

\bibitem{kurakin2018adversarial}
Alexey Kurakin, Ian Goodfellow, Samy Bengio, Yinpeng Dong, Fangzhou Liao, Ming
  Liang, Tianyu Pang, Jun Zhu, Xiaolin Hu, Cihang Xie, et~al.
\newblock Adversarial attacks and defences competition.
\newblock In {\em The NIPS'17 competition: building intelligent systems}, pages
  195--231. Springer, 2018.

\bibitem{li2015visual}
Guanbin Li and Yizhou Yu.
\newblock Visual saliency based on multiscale deep features.
\newblock In {\em CVPR}, pages 5455--5463, 2015.

\bibitem{li2020backdoor}
Yiming Li, Baoyuan Wu, Yong Jiang, Zhifeng Li, and Shu-Tao Xia.
\newblock Backdoor learning: A survey.
\newblock {\em arXiv preprint arXiv:2007.08745}, 2020.

\bibitem{li2020rethinking}
Yiming Li, Tongqing Zhai, Baoyuan Wu, Yong Jiang, Zhifeng Li, and Shutao Xia.
\newblock Rethinking the trigger of backdoor attack.
\newblock {\em arXiv preprint arXiv:2004.04692}, 2020.

\bibitem{liu2018fine}
Kang Liu, Brendan Dolan-Gavitt, and Siddharth Garg.
\newblock Fine-pruning: Defending against backdooring attacks on deep neural
  networks.
\newblock In {\em International symposium on research in attacks, intrusions,
  and defenses}, pages 273--294. Springer, 2018.

\bibitem{liu2017trojaning}
Yingqi Liu, Shiqing Ma, Yousra Aafer, Wen-Chuan Lee, Juan Zhai, Weihang Wang,
  and Xiangyu Zhang.
\newblock Trojaning attack on neural networks.
\newblock 2017.

\bibitem{liu2017neural}
Yuntao Liu, Yang Xie, and Ankur Srivastava.
\newblock Neural trojans.
\newblock In {\em International conference on computer design}, pages 45--48,
  2017.

\bibitem{lundberg2017unified}
Scott~M Lundberg and Su-In Lee.
\newblock A unified approach to interpreting model predictions.
\newblock In {\em Neural information processing systems}, pages 4765--4774,
  2017.

\bibitem{moosavi2017universal}
Seyed-Mohsen Moosavi-Dezfooli, Alhussein Fawzi, Omar Fawzi, and Pascal
  Frossard.
\newblock Universal adversarial perturbations.
\newblock In {\em Conference on computer vision and pattern recognition}, pages
  1765--1773, 2017.

\bibitem{mopuri2018generalizable}
Konda~Reddy Mopuri, Aditya Ganeshan, and R~Venkatesh Babu.
\newblock Generalizable data-free objective for crafting universal adversarial
  perturbations.
\newblock {\em IEEE transactions on pattern analysis and machine intelligence},
  41(10):2452--2465, 2018.

\bibitem{olah2020zoom}
Chris Olah, Nick Cammarata, Ludwig Schubert, Gabriel Goh, Michael Petrov, and
  Shan Carter.
\newblock Zoom in: An introduction to circuits.
\newblock {\em Distill}, 5(3):e00024--001, 2020.

\bibitem{qi2017pointnet}
Charles~R Qi, Hao Su, Kaichun Mo, and Leonidas~J Guibas.
\newblock Pointnet: Deep learning on point sets for 3d classification and
  segmentation.
\newblock In {\em Conference on computer vision and pattern recognition}, pages
  652--660, 2017.

\bibitem{qiao2019defending}
Ximing Qiao, Yukun Yang, and Hai Li.
\newblock Defending neural backdoors via generative distribution modeling.
\newblock In {\em Neural information processing systems}, pages 14004--14013,
  2019.

\bibitem{rakin2020tbt}
Adnan~Siraj Rakin, Zhezhi He, and Deliang Fan.
\newblock Tbt: Targeted neural network attack with bit trojan.
\newblock In {\em Conference on computer vision and pattern recognition}, pages
  13198--13207, 2020.

\bibitem{schwarzschild2020just}
Avi Schwarzschild, Micah Goldblum, Arjun Gupta, John~P Dickerson, and Tom
  Goldstein.
\newblock Just how toxic is data poisoning? a unified benchmark for backdoor
  and data poisoning attacks.
\newblock {\em arXiv preprint arXiv:2006.12557}, 2020.

\bibitem{selvaraju2017grad}
Ramprasaath~R Selvaraju, Michael Cogswell, Abhishek Das, Ramakrishna Vedantam,
  Devi Parikh, and Dhruv Batra.
\newblock Grad-cam: Visual explanations from deep networks via gradient-based
  localization.
\newblock In {\em Conference on computer vision and pattern recognition}, pages
  618--626, 2017.

\bibitem{shokri2015privacy}
Reza Shokri and Vitaly Shmatikov.
\newblock Privacy-preserving deep learning.
\newblock In {\em Proceedings of the 22nd ACM SIGSAC conference on computer and
  communications security}, pages 1310--1321, 2015.

\bibitem{simonyan2013deep}
Karen Simonyan, Andrea Vedaldi, and Andrew Zisserman.
\newblock Deep inside convolutional networks: Visualising image classification
  models and saliency maps.
\newblock {\em arXiv preprint arXiv:1312.6034}, 2013.

\bibitem{sundararajan2017axiomatic}
Mukund Sundararajan, Ankur Taly, and Qiqi Yan.
\newblock Axiomatic attribution for deep networks.
\newblock {\em arXiv preprint arXiv:1703.01365}, 2017.

\bibitem{szegedy2013intriguing}
Christian Szegedy, Wojciech Zaremba, Ilya Sutskever, Joan Bruna, Dumitru Erhan,
  Ian Goodfellow, and Rob Fergus.
\newblock Intriguing properties of neural networks.
\newblock {\em arXiv preprint arXiv:1312.6199}, 2013.

\bibitem{tang2020embarrassingly}
Ruixiang Tang, Mengnan Du, Ninghao Liu, Fan Yang, and Xia Hu.
\newblock An embarrassingly simple approach for trojan attack in deep neural
  networks.
\newblock In {\em International conference on knowledge discovery \& Data
  Mining}, pages 218--228, 2020.

\bibitem{thys2019fooling}
Simen Thys, Wiebe Van~Ranst, and Toon Goedem{\'e}.
\newblock Fooling automated surveillance cameras: adversarial patches to attack
  person detection.
\newblock In {\em Conference on computer vision and pattern recognition}, pages
  0--0, 2019.

\bibitem{turner2019label}
Alexander Turner, Dimitris Tsipras, and Aleksander Madry.
\newblock Label-consistent backdoor attacks.
\newblock {\em arXiv preprint arXiv:1912.02771}, 2019.

\bibitem{vaswani2017attention}
Ashish Vaswani, Noam Shazeer, Niki Parmar, Jakob Uszkoreit, Llion Jones,
  Aidan~N Gomez, {\L}ukasz Kaiser, and Illia Polosukhin.
\newblock Attention is all you need.
\newblock In {\em Neural information processing systems}, pages 5998--6008,
  2017.

\bibitem{villarreal2020confoc}
Miguel Villarreal-Vasquez and Bharat Bhargava.
\newblock Confoc: Content-focus protection against trojan attacks on neural
  networks.
\newblock {\em arXiv preprint arXiv:2007.00711}, 2020.

\bibitem{wang2019neural}
Bolun Wang, Yuanshun Yao, Shawn Shan, Huiying Li, Bimal Viswanath, Haitao
  Zheng, and Ben~Y Zhao.
\newblock Neural cleanse: Identifying and mitigating backdoor attacks in neural
  networks.
\newblock In {\em 2019 IEEE Symposium on Security and Privacy (SP)}, pages
  707--723. IEEE, 2019.

\bibitem{wang2020scout}
Pei Wang and Nuno Vasconcelos.
\newblock Scout: Self-aware discriminant counterfactual explanations.
\newblock In {\em Conference on computer vision and pattern recognition}, pages
  8981--8990, 2020.

\bibitem{wang2020practical}
Ren Wang, Gaoyuan Zhang, Sijia Liu, Pin-Yu Chen, Jinjun Xiong, and Meng Wang.
\newblock Practical detection of trojan neural networks: Data-limited and
  data-free cases.
\newblock {\em arXiv preprint arXiv:2007.15802}, 2020.

\bibitem{wang2020backdoor}
Shuo Wang, Surya Nepal, Carsten Rudolph, Marthie Grobler, Shangyu Chen, and
  Tianle Chen.
\newblock Backdoor attacks against transfer learning with pre-trained deep
  learning models.
\newblock {\em arXiv preprint arXiv:2001.03274}, 2020.

\bibitem{wong2020fast}
Eric Wong, Leslie Rice, and J~Zico Kolter.
\newblock Fast is better than free: Revisiting adversarial training.
\newblock {\em arXiv preprint arXiv:2001.03994}, 2020.

\bibitem{yi2016lift}
Kwang~Moo Yi, Eduard Trulls, Vincent Lepetit, and Pascal Fua.
\newblock Lift: Learned invariant feature transform.
\newblock In {\em European conference on computer vision}, pages 467--483.
  Springer, 2016.

\bibitem{zaheer2017deep}
Manzil Zaheer, Satwik Kottur, Siamak Ravanbakhsh, Barnabas Poczos, Russ~R
  Salakhutdinov, and Alexander~J Smola.
\newblock Deep sets.
\newblock In {\em Neural information processing systems}, pages 3391--3401,
  2017.

\bibitem{zhang2020cassandra}
Xiaoyu Zhang, Ajmal Mian, Rohit Gupta, Nazanin Rahnavard, and Mubarak Shah.
\newblock Cassandra: Detecting trojaned networks from adversarial
  perturbations.
\newblock {\em arXiv preprint arXiv:2007.14433}, 2020.

\bibitem{zhang2020backdoor}
Zaixi Zhang, Jinyuan Jia, Binghui Wang, and Neil~Zhenqiang Gong.
\newblock Backdoor attacks to graph neural networks.
\newblock {\em arXiv preprint arXiv:2006.11165}, 2020.

\bibitem{zhao2020bridging}
Pu Zhao, Pin-Yu Chen, Payel Das, Karthikeyan~Natesan Ramamurthy, and Xue Lin.
\newblock Bridging mode connectivity in loss landscapes and adversarial
  robustness.
\newblock {\em arXiv preprint arXiv:2005.00060}, 2020.

\bibitem{zhu2020gangsweep}
Liuwan Zhu, Rui Ning, Cong Wang, Chunsheng Xin, and Hongyi Wu.
\newblock Gangsweep: Sweep out neural backdoors by gan.
\newblock In {\em International conference on multimedia}, pages 3173--3181,
  2020.

\end{thebibliography}
}

\newpage
\onecolumn

\section{Implementation Details}

We provide implementation details for our model and SOTA methods. 

\paragraph{Our model:} We compute the per-class attributions using Integrated Gradient \cite{sundararajan2017axiomatic} and Gradient$\times$Activation using the Captum library\cite{kokhlikyan2020captum}. 
The temporal encoder in \trinity\; is either a CNN or a Transformer.  
The CNN consists of two 1D-CNN layers with $\text{kernel\_size}=13$, $\text{stride}=1$, and $\text{output\_channels}=4$ and $16$. There is a 1D-MaxPool ($\text{kernel\_size=9}$, $\text{stride}=2$) and a ReLU after each layer. 
We use the Transformer encoder with positional encoding \cite{vaswani2017attention} and its hyperparameters such as $\text{number\_of\_heads}$ and $\text{num\_layers}$ are set using the performance on the validation split. 
We also use the validation split to set the learning rate and perform early stopping.    
To mitigate issues with small training datasets, we use an ensemble strategy by training $5$ models with different random seeds and average their output scores. We use the Adam optimizer with cross-entropy loss, and set batch-size to $32$ for optimization.

\paragraph{Cassandra \cite{zhang2020cassandra}:}  uses Universal Adversarial Perturbations (UAP) \cite{moosavi2017universal} based attack to generate a universal noise pattern from the clean images. 
The attack is untargeted and thus generates a single noise pattern for all the clean images, which would cause misclassification when added to the input.
Cassandra is based on the intuition that such noise pattern contains useful information about the decision boundaries, and would thus help differentiate between a 
benign and a Trojaned model. 
Cassandra generates such noise patterns for all the DNNs in a dataset and uses it as input to train a deep Trojan detector.
We implemented this model since the code was not publicly available. 

Within UAP we used the FGSM attack instead of the original DeepFool attack since the latter was too slow \cite{hao2020adversarial}. We used the ART library\footnote{\url{https://github.com/Trusted-AI/adversarial-robustness-toolbox}} to implement this attack with norm set to L$1$, eps=$5000$, delta=$0.25$, and the maximum iterations for UAP were set to $20$. 
We set these parameters based on extensive experimentation and the information provided in the paper. 
We also found this attack to be more effective than the $L_{\inf}$ as used in the paper. 
Following the paper, we used Mobilenet-v2 \cite{howard2017mobilenets} to encode the noise pattern, a $5$ layer MLP to encode the $50\times50$ maximum energy crop from the noise pattern, and also appended the 
$\frac{\text{L1 Norm}}{\text{Fooling Ratio}}$ to the outputs of the previous two networks. We only used a single branch based on the L$1$ attack as we found it give similar results. We use Adam optimizer, cross-entropy loss, and set batch-size to $32$ for optimization. The learning rate was set sing the validation set.    

\paragraph{Neural Cleanse \cite{wang2019neural}:} tries to reverse engineer the trigger using an optimization that generates a sparse mask and a trigger pattern that causes misclassification to a specific target class. 
We used the code\footnote{\url{https://github.com/bolunwang/backdoor}} provided by authors to implement this method. We followed the paper and repeated the optimization for each dataset class (as the possible target label) and obtained their $L1$ norms. 
We then identified the trigger and its associated class based on it showing up as an outlier (using Mean Absolute Deviation (MAD)) with the smaller $L1$ norm in the distribution. Each model was assigned a detection score (measuring if the model is Trojaned) of $1$ if a class was identified as an outlier and $0$ otherwise. We also experimented with using other measures such as using the anomaly index from MAD as detection scores but did not found them to be helpful.
We selected parameters such as attack accuracy, learning rate, regularization hyper-parameter using the validation set. 

\paragraph{Universal Litmus Patterns \cite{kolouri2020universal}:} learns a set of universal noise patterns that are passed through a model, pools the logit outputs, and then classify into a benign or a Trojaned model. We used the code\footnote{\url{https://github.com/UMBCvision/Universal-Litmus-Patterns/}} provided by the authors for implementation. We selected the parameters such as number of noise patterns and learning rate using the validation set. We report performance for only Triggered-MNIST and NIST-TrojAI-Round1 since the model was too slow to train on Round2 and Round3 due to the time required to load a DNN in the GPU for every iteration (each datapoint is a DNN during learning).

\section{Detailed for Triggered MNIST Dataset}
\begin{figure*}[!h]
	\begin{center}
        \includegraphics[width=0.6\textwidth]{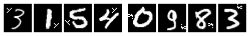}
	\end{center}
	\caption{Few poisoned samples from Triggered MNIST Dataset showing the two types of triggers used to generate the dataset}
	\label{image:mnist}
  \end{figure*}

We use the code provided by NIST\footnote{\url{https://github.com/trojai/trojai}} to generate $810$ DNNs ($50\%$ are Trojaned). To generate poisoned data, we  inserted two types of triggers into clean MNIST images as shown in \autoref{image:mnist}. We used $3$ model architectures ModdedBadNet (2 Conv + 1 Dense layers), BadNet (2 Conv + 2 Dense layers) and ModdedLeNet5 (3 Conv + 2 Dense layers) to generate the dataset. For more details on the poisoned data and model architecture refer to \cite{zhang2020cassandra}. The dataset comprises of $810$ models of which $405$ models are benign and are trained with random seeds with $\SI{98.59 \pm 0.94}{\percent}$ validation accuracy on clean data. 
For training Trojaned models we randomly selected combinations of model architecture, trigger type, and different trigger fractions (from $0.05, 0.1, 0.15, 0.2$). We only used any-to-one targeted attacks. 
We also randomly varied the number of source classes to make the dataset challenging ($len(C_s) \leq 2$ or $len(C_s) \geq 5$). 
We discarded models with less than $90\%$ validation accuracy on corresponding triggered data to keep only robustly trained Trojaned models. This remaining $405$ Trojaned models had $\SI{97.41 \pm 6.5}{\percent}$ validation accuracy on clean data and $\SI{97.07 \pm 3}{\percent}$ validation accuracy on corresponding triggered data. Finally we use $200$ clean images with $20$ images per class.

\end{document}